\documentclass[10pt,twocolumn,letterpaper]{article}

\usepackage[pagenumbers]{wacv} 

\usepackage{graphicx}
\usepackage{amsmath}
\usepackage{amssymb}
\usepackage{booktabs}
\usepackage[utf8]{inputenc} 
\usepackage[T1]{fontenc}    
\usepackage{url}            
\usepackage{amsfonts}       
\usepackage{nicefrac}       
\usepackage{microtype}      
\usepackage{xcolor}         
\usepackage{amsmath}
\usepackage{bbm}
\usepackage{multirow}
\usepackage{adjustbox}
\usepackage{subcaption}
\usepackage{algorithm} 
\usepackage{algorithmic}
\usepackage{setspace}
\usepackage[algo2e,ruled,vlined]{algorithm2e}

\usepackage{caption}

\newcommand{\cutsectionup}{\vspace{-4pt}}
\newcommand{\cutsectiondown}{\vspace{-6pt}}
\newcommand{\cutsubsection}{\vspace{-3pt}}

\usepackage[subtle]{savetrees}

\usepackage[pagebackref,breaklinks,colorlinks]{hyperref}

\usepackage[capitalize]{cleveref}
\crefname{section}{Sec.}{Secs.}
\Crefname{section}{Section}{Sections}
\Crefname{table}{Table}{Tables}
\crefname{table}{Tab.}{Tabs.}

\def\wacvPaperID{1731} 
\def\confName{WACV}
\def\confYear{2025}

\begin{document}

\title{Ego-VPA: Egocentric Video Understanding with Parameter-efficient Adaptation}

\author{Tz-Ying Wu$^{1,2}$ \quad Kyle Min$^1$ \quad Subarna Tripathi$^1$ \quad Nuno Vasconcelos$^2$\\[-1.5em]
\and
$^1$Intel Labs\\
{\tt\small\{tz-ying.wu,kyle.min,subarna.tripathi\}@intel.com}\\
\and
$^2$UC San Diego\\
{\tt\small nvasconcelos@ucsd.edu}
}
\maketitle

\begin{abstract}
  Video understanding typically requires fine-tuning the large backbone when adapting to new domains.
  In this paper, we leverage the egocentric video foundation models (Ego-VFMs) based on video-language pre-training and propose a parameter-efficient adaptation for egocentric video tasks, namely Ego-VPA.
  It employs a local sparse approximation for each video frame/text feature using the basis prompts, and the selected basis prompts are used to synthesize video/text prompts. 
  Since the basis prompts are shared across frames and modalities, it models context fusion and cross-modal transfer in an efficient fashion. Experiments show that Ego-VPA excels in lightweight adaptation (with only $0.84\%$ learnable parameters), largely improving over baselines and reaching the performance of full fine-tuning.
\end{abstract}

\cutsectionup
\section{Introduction}\label{sec:intro}
\cutsectiondown
Video understanding models have achieved satisfactory performance on various downstream tasks, such as video captioning~\cite{sun2019videobert,seo2022end,yang2023vid2seq}, retrieval~\cite{bain2021frozen,miech2020end} and action classification~\cite{miech2019howto100m,zhao2023lavila}. These models are typically trained on the supervised video datasets of interest~\cite{caba2015activitynet,carreira2017quo,gu2018ava}. 
Inspired by visual-language contrastive learning~\cite{oord2018representation,miech2020end}, recent research has been shifted to training video foundation models (VFMs)~\cite{bain2021frozen,kevin2022egovlp,pramanick2023egovlpv2,ashutosh2023hiervl,yang2023vid2seq,zhao2023lavila} on large datasets, 
to produce representations that generalize to multiple tasks.
Prior work has focused on aligning the video and text representations of the VFM, by developing novel training objectives ~\cite{xu2021videoclip,kevin2022egovlp,ashutosh2023hiervl}, or leveraging data from other modalities, such as speech recognition~\cite{sun2019videobert} or language models~\cite{zhao2023lavila}. While this research has improved zero-shot performance on unseen domains, there exists a gap between the latter and that of full VFM fine-tuning~\cite{wang2023allinone,kevin2022egovlp,zhao2023lavila}. This gap ensues from a statistics mismatch between the pretraining data and the application of interest, due to factors such as background variability, video context, etc. It reduces the practical value of VFMs, since fine-tuning usually requires extensive computation, and model parameters can grow exponentially when adapting to multiple tasks.
This work focuses on the lightweight adaptation of VFMs for egocentric (first-person view) videos. 

\begin{figure}[t!]
    \centering
        \resizebox{\linewidth}{!}{
        \begin{minipage}[t]{0.47\linewidth}
            \hspace{-10pt}
            \includegraphics[width=\linewidth]{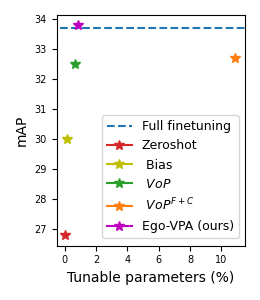} 
        \end{minipage}
        \hspace{-10pt}
        \begin{minipage}[t]{0.55\linewidth}
            \hspace{-10pt}
            \includegraphics[width=\linewidth]{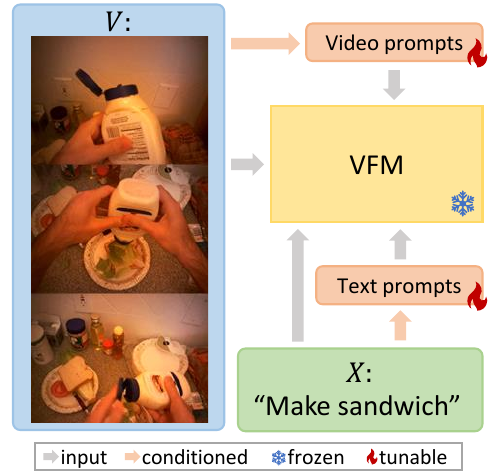}
        \end{minipage}
        }
        \vspace{-18pt}
        \captionof{figure}{Ego-VPA leverages context-aware prompts to achieve parameter-efficient adaptation for egocentric videos. (Left) Performance vs tunable parameters; (Right) Cross modality prompt-tuning in Ego-VPA where the VFM is frozen.}
        \label{fig:teaser}
\end{figure}

The gap between zero-shot and fine-tuning performance also holds when image-based foundation models (IFMs) (i.e.  CLIP~\cite{clip}) are used for image recognition~\cite{wortsman2022robust}. 
Several parameter-efficient adaptation techniques have been developed, including the use of adapters~\cite{CLIPAdapter} and prompt-tuning~\cite{coop,cocoop,jia2022visual}.
This inspired a few recent applications to video understanding~\cite{huang2023vop,wasim2023vita,sung2022vl,pan2022st}, based on the use of IFMs to encode individual video frames. We hypothesize that this is insufficient for egocentric video, since IFMs are pretrained on third-person views, not first-person, and lack temporal reasoning capabilities. The use of existing lightweight adaptation techniques is insufficient to overcome these barriers. We validate this hypothesis by demonstrating the inefficiency of two IFM-based video models~\cite{ni2022expanding, wasim2023vita} for egocentric video understanding, as shown in Table~\ref{table:clip_vfm}.

\begin{figure}[t!]
    \centering
    \begin{minipage}[h]{\linewidth}
        \centering
        \begin{minipage}[h]{0.51\linewidth}
        \vspace{-8pt}
        \setlength{\tabcolsep}{2pt}
        \resizebox{\linewidth}{!}{
        \begin{tabular}{l|c|c}
        \toprule
        Method & zero-shot & fine-tuned\\ \midrule
        {\it CLIP-based VFM} & & \\
        ~~X-CLIP~\cite{ni2022expanding}    & $24.0$ & $30.0^{\star}$ \\
        ~~Vita-CLIP~\cite{wasim2023vita}   & $25.8$ & $31.3^{\star}$ \\ \midrule
        {\it Egocentric VFM} & & \\
        ~~LaViLa~\cite{zhao2023lavila} & $26.8$ & $33.7~$\\ \bottomrule
        \end{tabular}
        }
        \end{minipage}
        \hfill
        \begin{minipage}[h]{0.48\linewidth}
        \captionof{table}{The zero-shot - fine-tuned performance gap (mAP on Charades-Ego) exists in both CLIP-based VFMs~\cite{ni2022expanding,wasim2023vita} and Ego-VFMs~\cite{zhao2023lavila} . $^{\star}$ denotes that only the prompts/adapters are fine-tuned on Charades-Ego.}\label{table:clip_vfm}
        \end{minipage}
    \end{minipage}
    \vspace{-4pt}
\end{figure}

In this work, we investigate efficient prompt tuning based frameworks to address light-weight adaptation of VFMs, in particular Ego-VFMs~\cite{kevin2022egovlp,pramanick2023egovlpv2,zhao2023lavila} which are pretrained on egocentric videos. Since no prior work investigated this problem, we begin with introducing several baselines, including adding learnable prompts only to the text or the video encoder. Single-modality prompts unsurprisingly lead to marginal improvements over zero-shot VFM performance as they fail to capture connections between the text and spatial-temporal context. We then show that improved results can be obtained by prompt-tuning both the video and text encoders, using the recent VoP approach~\cite{huang2023vop}, which uses an additional module, a bi-directional LSTM, to connect visual prompts across frames, according to context. While achieving good performance, this introduces a large parameter overhead, compromising the lightweight nature of the adaptation (See trainable parameters in Table~\ref{table:main}).

To address this, we propose a parameter-efficient prompt-tuning approach, {\bf Ego-VPA}, for Ego-VFMs. Ego-VPA uses an encoder to project frame feature vectors into a latent prompt space. It then learns an orthogonal basis for this space, the {\it prompt basis}, which can be seen as a principal component analysis of prompt space. Given the latent space projection of a frame feature vector, Ego-VPA then determines the subspace of $k$ basis-prompts that best approximates it, in the least squares sense. The selected basis prompts are then used to synthesize $k$ video prompts for the video frame, using a linear latent decoder. Since it is trained over the entire dataset, the prompt basis is representative of all frames, allowing efficient cross-frame context modeling. In addition, since the basis prompts capture the semantic content of the video, the method can be naturally extended to cross-modal prompt synthesis, where the basis prompts that best reconstruct the projected video/text feature are used to synthesize video/text prompts. Figure~\ref{fig:teaser} summarizes the overall cross-modal prompt-tuning.

To highlight the effectiveness and efficiency of Ego-VPA, three popular egocentric video datasets (i.e. Charades-Ego~\cite{sigurdsson2018charades}, EGTEA~\cite{li2018eye}, and EPIC-Kitchens-100~\cite{damen2022rescaling}) are evaluated.
We show that Ego-VPA outperforms the prompt-tuning baselines and is even superior to the fully fine-tuned Ego-VFM on Charades-Ego and EGTEA. More importantly, Ego-VPA only requires 0.84\% additional model parameters, which is much more efficient than other baselines (See Figure~\ref{fig:teaser}). Ablations show that Ego-VPA consistently outperforms the baselines when different numbers of frames per video or amounts of training data are used.  

Overall, we make three contributions to the efficient adaptation for egocentric video understanding.
First, we show that prompt-tuning video encoders based on IFMs are suboptimal for egocentric video tasks due to the inherent domain gap.
Second, we propose several baselines that prompt-tune the existing Ego-VFMs and show that these baselines are not effective and efficient. Finally, we propose a novel and efficient prompt-tuning approach, Ego-VPA, that utilizes a local subspace approximation with shared basis prompts for cross-modal prompt synthesis, enabling context reasoning across frames and modalities.

\cutsectionup
\section{Related Works}\label{sec:rw}
\cutsectiondown
\noindent\textbf{Video foundation models (VFMs).}
VFM learns a generalizable representation for videos, which is applicable to many downstream tasks, such as video captioning~\cite{sun2019videobert,seo2022end,yang2023vid2seq}, retrieval~\cite{bain2021frozen,miech2020end}, action classification~\cite{miech2019howto100m,ashutosh2023hiervl,zhao2023lavila}. One approach to achieve this is to expand existing image-language foundation models (IFMs)~\cite{clip} to the video domain~\cite{lei2021less,wang2021actionclip,lin2022frozen,xue2022clip,ni2022expanding,rasheed2023fine,wasim2023vita,huang2023vop}. These works show promising results on some short-term or third-person-view video understanding tasks~\cite{kay2017kinetics,kuehne2011hmdb,soomro2012ucf101}. However, since most of the existing IFMs~\cite{clip} are pretrained on static internet images, IFMs require additional temporal reasoning modules to capture scene dynamics and may fail to extract meaningful representations for frames in egocentric videos.
Another line of works adopts video-language pre-training (VLP)~\cite{luo2020univl,bain2021frozen,xu2021videoclip,zellers2021merlot,seo2022end,kevin2022egovlp,pramanick2023egovlpv2,yang2023vid2seq,wang2023allinone,ashutosh2023hiervl,li2023lavender,zhao2023lavila} to learn transferable spatial-temporal representation from large-scale video datasets~\cite{bain2021frozen,miech2019howto100m,grauman2022ego4d}. 
For example, UniVL~\cite{luo2020univl}, All-in-one~\cite{wang2023allinone} and Lavender~\cite{li2023lavender}
propose a general pre-training method that can support many tasks and achieve solid zero-shot performance. Recently, LaViLa~\cite{zhao2023lavila} shows that VLP can benefit from the dense narrations generated by Large Language Models (LLMs).
In this work, we focus on LaViLa, which is the SOTA Ego-VFM 
on egocentric videos~\cite{sigurdsson2018charades,li2018eye,damen2022rescaling}.

\noindent\textbf{Parameter-efficient adaptation.}
Adaptation techniques have been widely used in natural language processing (NLP) for efficiently adapt pretrained LLMs for domain-specific tasks. For example, task-specific modules (i.e. adapters)~\cite{pmlr-v97-stickland19a,pmlr-v97-houlsby19a,houlsby2019parameter,pfeiffer2020adapterhub} are integrated into transformers for efficient adaptation. An alternative is to prompt-tune~\cite{li2021prefix,shin2020autoprompt} the large models, where extra tokens are prepended to the model input and are optimized with domain-specific losses. In both cases, the large model remains fixed. These adaptation techniques in NLP are then adopted in the computer vision field. For example, VL-adapter~\cite{sung2022vl} and ST-adapter~\cite{pan2022st} propose effective adapter-based methods for the tasks of image-language and video understanding, respectively.  VPT~\cite{jia2022visual}, CoOp~\cite{coop} and CoCoOp~\cite{cocoop} prompt-tune image transformers and CLIP~\cite{clip} for image recognition tasks. 
Recently, VoP~\cite{huang2023vop} extended prompt-tuning techniques to the video-language domain by applying video and text prompts. We adapt VoP and its variants to Ego-VFM as strong baselines and propose an efficient way to model context fusion atop the baselines that allows knowledge sharing across frames and modalities.

\cutsectionup
\section{Egocentric Video Understanding with VFMs}
\cutsubsection
\subsection{VFM Preliminaries}\label{sec:prelim}
\cutsubsection
Inspired by the success of IFMs on image applications, two types of VFMs have been proposed for video. One extends existing IFMs (e.g. CLIP) to the video domain~\cite{ni2022expanding,wasim2023vita,huang2023vop}, by first encoding individual frames with an IFM and then fusing them with a temporal reasoning component. The other directly employs a video encoder to learn spatial-temporal representations~\cite{wang2023allinone,zhao2023lavila}.
These models use video-language pretraining to learn representations that align two modalities.
Similar to IFMs, most VFMs~\cite{bain2021frozen,ashutosh2023hiervl,zhao2023lavila} adopt dual-encoder design, where a video encoder $\phi_{vid}$ and a text encoder $\phi_{txt}$ extract features from each video $V$ and its corresponding text description $X$ (e.g., ``turning on the light"), respectively.
$\phi_{vid}$ and $\phi_{txt}$ are usually transformers~\cite{bain2021frozen,ashutosh2023hiervl,zhao2023lavila}.
Given an input sequence ${\bf Z}^{(0)}\in\mathbb{R}^{N_t\times d}$ with $N_t$ tokens, the $L$-layer transformer performs the mapping
\begin{equation}
    {\bf Z}^{(l+1)} = Att({\bf Z}^{(l)})~_{l=0...{L-1}}\label{eq:transformer},
\end{equation}
where $Att(\cdot)$ is a transformer block~\cite{vaswani2017attention}. For $l=0$, a learnable positional encoding that specifies the relative position between the tokens is added.
In the following, we omit all learnable positional encoding for brevity and use the subscript $vid$ and $txt$ to represent variables from the video and text domains, respectively.

\noindent\textbf{Text encoder $\phi_{txt}$.}
Given a textual description $X$, each tokenized word is mapped into a text embedding ${\bf z}^{(0)}_i\in\mathbb{R}^{d_{txt}}$.
The transformer of Eq.~(\ref{eq:transformer}) takes ${\bf Z}_{txt}^{(0)} = ({\bf z}^{(0)}_{[SOS]},{\bf z}^{(0)}_1,...,{\bf z}^{(0)}_{N_w},{\bf z}^{(0)}_{[EOS]})^T$ as the input, where ${\bf z}^{(l)}_{[SOS]},{\bf z}^{(l)}_{[EOS]}\in\mathbb{R}^{d_{txt}}$ are special tokens representing the start and the end of the sequence, which is mapped into a token sequence ${\bf Z}_{txt}^{(l)}$ at each transformer layer $l$, and defines the text feature as $\phi_{txt}(X) = {\bf z}^{(L)}_{[EOS]}$, where $L$ is the last transformer layer.

\noindent\textbf{Video encoder $\phi_{vid}$.}
Given a video $V$ with $T$ frames, each frame ${\bf V}^f$ is mapped into patch embeddings $\{{\bf z}_{(f,p)}^{(0)}\}_{p=1}^{N_p}$, where $N_p$ is the number of patches.
A [CLS] token ${\bf z}_{[CLS]}^{(0)}\in\mathbb{R}^{d_{vid}}$ is prepended to the input sequence, i.e. ${\bf Z}^{(0)}_{vid}=({\bf z}_{[CLS]}^{(0)}, {\bf z}_{(1,1)}^{(0)}, ...,{\bf z}_{(T,N_p)}^{(0)})^T\in\mathbb{R}^{(N_p T+1)\times d_{vid}}$.
The video feature $\phi_{vid}(V)={\bf z}^{(L)}_{[CLS]}$ is then extracted with the video encoder $\phi_{vid}$ with transformer mapping of Eq.~(\ref{eq:transformer}).
Since the $(N_p T+1)$ attention computations of each patch induce large memory overhead, there is a need to trade-off between space and time. VFMs typically use the TimeSformer~\cite{gberta_2021_ICML} architecture, which utilizes two types of attention: a spatial attention block that only attends to features from the same time/frame (i.e. $({\bf z}_{[CLS]},{\bf z}_{(f,1)},...,{\bf z}_{(f,N_p)})$); and a time attention block that only attends to features from the same location (i.e., $({\bf z}_{[CLS]},{\bf z}_{(1,p)},...,{\bf z}_{(T,p)})$). The two attention blocks are interleaved to construct each transformer block of Eq.~(\ref{eq:transformer}), reducing the attention computations per patch to $(N_p+T+2)$. This enables increasing the sampling rate of each video (e.g., from 4 frames in~\cite{bain2021frozen} to 16 frames). Note that this is especially important for egocentric videos as camera tends to move faster.

\noindent\textbf{Optimization.}
Following standard visual-language contrastive learning~\cite{clip} practice, most VFMs~\cite{bain2021frozen,ashutosh2023hiervl,zhao2023lavila} align the two modalities by optimizing the InfoNCE loss~\cite{oord2018representation}.
Given a video dataset $\mathcal{D}=\{V_i, X_i\}_{i=1}^N$, the loss for each batch $\mathcal{B}$ is
\begin{equation}
    {\cal L}_{cl} = -\frac{1}{|\mathcal{B}|}\left[\sum_{i\in\mathcal{B}} \log\frac{e^{({\bf v}_i^T {\bf t}_i/\tau)}}{\sum\limits_{j\in\mathcal{B}} e^{({\bf v}_i^T {\bf t}_j/\tau)}} +\log\frac{e^{({\bf t}_i^T {\bf v}_i/\tau)}}{\sum\limits_{j\in\mathcal{B}} e^{({\bf t}_i^T {\bf v}_j/\tau)}}\right],\label{eq:nce_loss}
\end{equation}
where $\tau$ is the temperature, ${\bf v}=\phi_{vid}(V)$ and ${\bf t}=\phi_{txt}(X)$.

\cutsubsection
\subsection{Generalization Ability of VFMs for Egocentric Video}
\cutsubsection
Standard VFMs are trained from large datasets, typically collected on the web, which consist mostly of exocentric (third-person-view) videos. We refer to them as Exo-VFMs. Egocentric videos are captured from a first-person view by wearable devices, which have many hand-object interactions and motion blurs caused by head and body movements. This renders Exo-VFMs sub-optimal for egocentric video understanding~\cite{kevin2022egovlp}. The introduction of the large-scale egocentric video dataset Ego4D~\cite{grauman2022ego4d} sparked the development of egocentric VFMs (Ego-VFMs)~\cite{kevin2022egovlp,pramanick2023egovlpv2,ashutosh2023hiervl,zhao2023lavila}.
However, while Ego4D is a very large dataset by egocentric video standards, it is unclear whether models learned from it can generalize to a broad set of egocentric video domains, namely to egocentric datasets other than Ego4D. To test this premise, we start by conducting some preliminary experiments on the generalization ability of Ego-VFMs to the popular Charades-Ego~\cite{sigurdsson2018charades} dataset. 

Table~\ref{table:clip_vfm} compares the performance of Ego-VFM LaViLa~\cite{zhao2023lavila} to that of two CLIP-based Exo-VFMs, X-CLIP~\cite{ni2022expanding} and Vita-CLIP~\cite{wasim2023vita}, under the zero-shot setting. While the Ego-VFM has improved performance, likely because it mitigates the change of perspective, the results are not drastically superior. In fact, there is a considerable gap between the zero-shot performance of the Ego-VFM and that after it is fine-tuned to Charades-Ego. The advantage of the Exo-VFMs is that they basically expand CLIP into a VFM using prompting. 
Since the IFM is fixed, this is a lightweight operation. For these models, the adaptation to new egovideo datsets is not difficult. As shown in the right column of Table~\ref{table:clip_vfm}, when their prompts and additional modules (beyond CLIP) are fine-tuned on Charades-Ego, their performance outperforms the zero-shot application of the Ego-VFM. Hence, from a practical standpoint, the adaptation of the Exo-VFMs to a new egocentric setting is superior to the zero-shot application of the Ego-VFM. However, they still underperform the fine-tuning performance of the latter. 

These observations suggest the following conclusions. First, Exo-VFMs like X-CLIP~\cite{ni2022expanding} and Vita-CLIP~\cite{wasim2023vita} are quite versatile and can cover large domain gaps with lightweight adaptation. However, they cannot match the performance of Ego-VFMs fine-tuned on the egovideo dataset of interest. Second, current Ego-VFMs appear to overfit to Ego4D dataset, requiring fine-tuning for effective performance on alternative egovideo datasets, like Charades-Ego. However, this requires a large computation and parameter overhead, and reduces the practical value of VFMs, especially if fine-tuning is required for multiple tasks.

This raises the question of how to design lightweight adaptation modules for the Ego-VFM models that, similarly to the Exo-VFMs of Table~\ref{table:clip_vfm}, can mitigate the performance gap across egovideo domains with reduced memory and training. We focus on the LaViLa~\cite{zhao2023lavila} model, since it is the current state-of-the-art Ego-VFM, where TimeSformer~\cite{gberta_2021_ICML} is adopted as the video encoder to trade off the space-time resolutions. In view of the established efficacy of prompt-tuning for other large foundation models~\cite{coop,jia2022visual}, we explore prompt-tuning as a parameter-efficient way to adapt Ego-VFMs to downstream egocentric video applications.

\begin{figure*}[t!]
    \centering
    \includegraphics[width=0.9\linewidth]{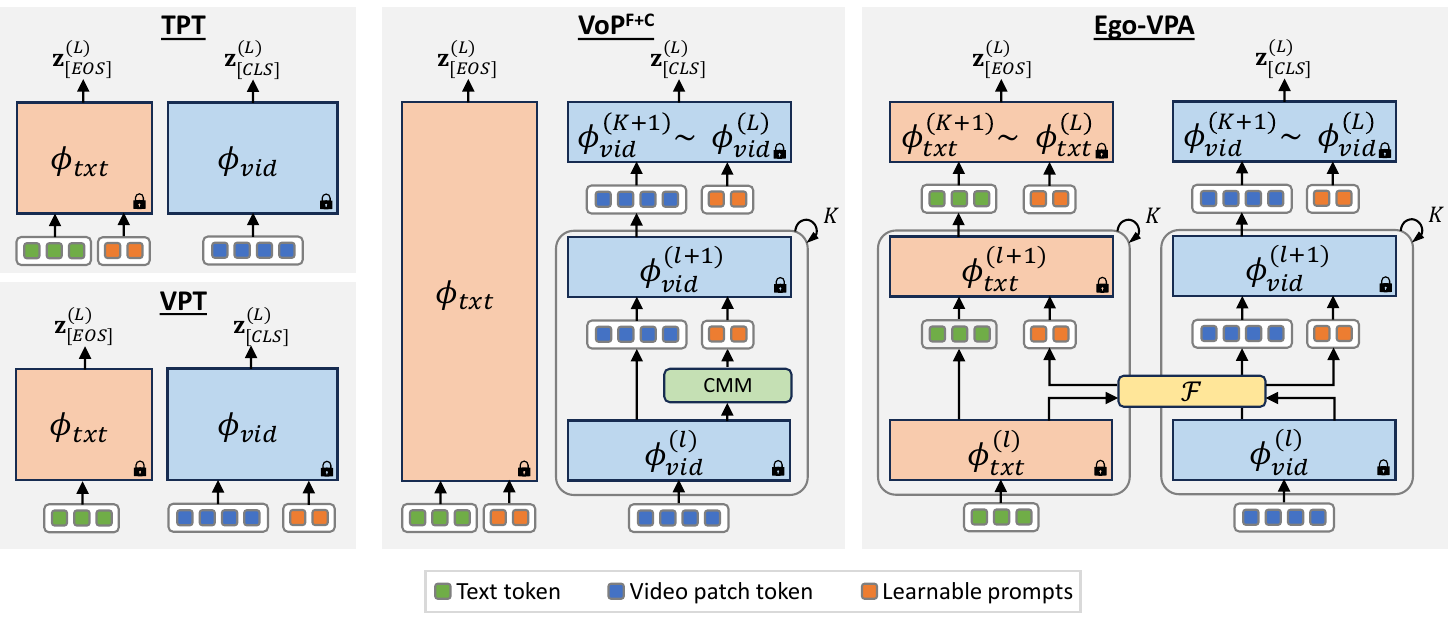}\vspace{-12pt}
    \caption{{\bf Models.} We adapt SOTA prompt-tuning methods to Ego-VFMs (See section~\ref{sec:settings}), i.e. TPT, VPT, and VoP\textsuperscript{F+C}, where CMM is a context modeling module. The proposed Ego-VPA leverages a set of basis prompts ${\cal F}$ for cross-modal prompt synthesis, enabling context modeling across frames and modalities in a highly efficient way (See section~\ref{sec:method}).
    }
    \label{fig:models}
\end{figure*}

\cutsectionup
\section{Ego-VFM Prompt-tuning Baselines}\label{sec:settings}
\cutsectiondown
Prompt-tuning introduces a set of learnable prompts, which are the only parameters optimized during the adaptation, leaving the rest of the VFM frozen. Specifically, the input sequence ${\bf Z}^{(0)}$ of the transformer of Eq.~(\ref{eq:transformer}) is augmented with learnable prompts ${\bf P}=({\bf p}_1,...,{\bf p}_M)\in\mathbb{R}^{d\times M}$.
While Exo-VFMs like X-CLIP~\cite{ni2022expanding} and Vita-CLIP~\cite{wasim2023vita} rely on prompt-tuning of IFMs, it is non-trivial to prompt-tune a TimeSformer-based VFM (such as LaViLa) as it utilizes divided space-time attention. We next introduce several baseline solutions, based on prompt-tuning methods in the literature. As shown in Figure~\ref{fig:models}, these insert prompts in the text encoder (TPT), video encoder (VPT), or both (VoP).  

\noindent\textbf{Text prompt-tuning (TPT).} Given text embedding ${\bf Z}_{txt}^{(0)}\in\mathbb{R}^{(N_w+2)\times d_{txt}}$, TPT prepends text prompts ${\bf P}_t=({\bf p}_{t,1},...,{\bf p}_{t,M_t})\in\mathbb{R}^{d_{txt}\times M_t}$ to the input text emebeddings, i.e. $\tilde{\bf Z}_{txt}^{(0)}=({\bf z}^{(0)}_{[SOS]},{\bf P}_t,{\bf z}^{(0)}_1,...,{\bf z}^{(0)}_{N_w},{\bf z}^{(0)}_{[EOS]})^T$.
The output feature $\phi_{txt}'(T)=\tilde{\bf z}^{(L)}_{[EOS]}$ is then extracted with the transformer in Eq.~(\ref{eq:transformer}).

\noindent\textbf{Video prompt-tuning (VPT).}
Given a video $V$, the video embedding of ${\bf Z}_{vid}^{(0)}\in\mathbb{R}^{(N_pT+1)\times d_{vid}}$ is extracted as described in section~\ref{sec:prelim}. To prompt-tune the video encoder, the visual prompts ${\bf P}_v=({\bf p}_{v,1},...,{\bf p}_{v,M_v})\in\mathbb{R}^{d_{vid}\times M_v}$ are prepended to the input sequence, i.e. $\tilde{\bf Z}_{vid}^{(0)}=({\bf z}^{(0)}_{[CLS]},{\bf P}_v, {\bf z}_{(1,1)}^{(0)}, ...,{\bf z}_{(T,N_p)}^{(0)})^T$.
Note that we apply prompt tuning only to the spatial attention blocks of the TimeSformer~\cite{gberta_2021_ICML} as we found that prompt-tuning both blocks has no additional gain (See Appendix).
Inspired by~\cite{jia2022visual}, beyond introducing visual prompts at the input layer, we further prepend layer-specific prompts ${\bf P}_v^{(l)}$ to every transformer layer. In the following, we omit the layer $l$ superscript of these deep prompts, for brevity. The output of VPT is then written as $\phi_{vid}'(V)=\tilde{\bf z}^{(L)}_{[CLS]}$.

\noindent\textbf{Text-video prompt-tuning (VoP).}
Recently, VoP~\cite{huang2023vop} expanded CLIP into a video model by prompt-tuning of both modalities. Several prompt-tuning variants are proposed in \cite{huang2023vop}. Vanilla VoP adopts both TPT and VPT, and the visual prompts are shared across all frames.
To propagate contextual information across frames, $VoP^{C}$ uses a context modeling module (CMM) to generate frame-specific prompts $\{{\bf P}_v^1,...,{\bf P}_v^T\}$ conditioned on the context information from other frames. To adapt it to LaViLa, while \cite{huang2023vop} adopts the [CLS] token from each frame-specific CLIP, we use ${\bf z}_f=AvgPool({\bf z}_{(f,1)},...,{\bf z}_{(f,N_p)})$ as the contextual feature of frame $f$.
In addition, we devise two types of prompting for the space attention block of the TimeSformer. 
For intra-frame attention, each patch token ${\bf z}_{(f,p)}$ can only attend to the prompts associated with frame $f$ (i.e. $({\bf z}_{[CLS]},{\bf P}_v^f,{\bf z}_{(f,1)},...,{\bf z}_{(f,N_p)})$). For inter-frame attention, each patch token ${\bf z}_{(f,p)}$ can attend to all the prompts across frames (i.e. $({\bf z}_{[CLS]},{\bf P}_v^1,...,{\bf P}_v^T,{\bf z}_{(f,1)},...,{\bf z}_{(f,N_p)})$). Following \cite{huang2023vop}, we adopt {\it intra-frame}/{\it inter-frame} attention in the first $K$/last $L-K$ transformer layers and integrate this strategy with $VoP^{C}$ as the $VoP^{F+C}$ variant.

\cutsectionup
\section{Ego-VPA}\label{sec:method}
\cutsectiondown
Our experiments (see Section~\ref{sec:sota}) show that $VoP^{F+C}$ prompting significantly improves the performance of the Ego-VFM on new egocentric datasets.  However, this method still falls short in two aspects. First, the CMM module, which is itself a bi-directional LSTM network, still requires a substantial number of model parameters. Second, there is no connection between text and visual prompts, limiting knowledge transfer across modalities. We next propose a novel prompt-tuning technique for Ego-VFMs such as LaViLa, denoted as Ego-VPA, which achieves context fusion across frames (like CMM) and modalities in an extremely lightweight fashion. This is implemented by the proposed prompt synthesis scheme, using a shared prompt basis. Note that we keep the inter-frame attention layers (last $L-K$ layers) identical to $VoP^{F+C}$ for fair comparison. 

\cutsubsection
\subsection{Video Prompt Synthesis}\label{sec:ps}
\cutsubsection
The main challenge for video prompting is how to design prompts that capture contextual connections across frames. Prompt design is almost trivial for stand-alone prompts, which are vectors with few parameters learned by back-propagation. However, once a prompt depends on other prompts, there is the need to learn the {\it functional dependence\/} between them. This can be done by learning a prompt synthesis network, e.g., a transformer that simultaneously generates prompts for image and text~\cite{zang2022unified}, or a recursive network, such as the LSTM of~\cite{huang2023vop} to synthesize prompts recursively, which is more sensible for video. While small, these networks have many more parameters than the prompts themselves, and can sacrifice the lightweight nature of the adaptation. For example, in Table~\ref{table:main}, both $VoP^{C}$ and  $VoP^{F+C}$ require about $10\%$ of the VFM parameters. 

\begin{figure}[t!]
    \centering
        \includegraphics[width=0.75\linewidth]{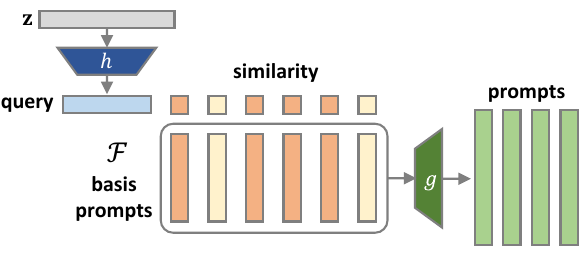}
        \vspace{-9pt}
        \caption{{\bf Prompt Synthesis.} Token ${\bf z}$ is projected into the subspace by $h(\cdot)$, and sparsely approximated by the top-$k$ similar prompts in the prompt basis ${\cal F}$, which are finally mapped into $k$ prompts by the mapping $g(\cdot)$. $(h,g)$ can be  $(h_{vid},g_{vid})$ or $(h_{txt},g_{txt})$ for visual or text prompt generation, respectively. $k=4$ in this illustration.
        }
        \label{fig:bank}
\end{figure}

\noindent\textbf{Frame-specific prompt synthesis.} In this work, we seek a more efficient solution, inspired by the compression literature and illustrated in Figure~\ref{fig:bank}. We assume that the prompt information lies on a lower dimensional latent space $\cal H$, onto which frame features ${\bf z}_f$ are mapped by an encoder $h_{vid}(\cdot): \mathbb{R}^{d_{vid}} \to {\cal H} \subset \mathbb{R}^{d_f}$, where $d_f \leq d_{vid}$. Prompts then inhabit a $B$-dimensional subspace ${\cal P}$ of $\cal H$ ($B < d_f$) of orthogonal basis ${\cal F}=\{{\bf f}_1,...,{\bf f}_{B}\}$. These vectors,  denoted as {\it basis prompts\/}, can be thought of as the principal components of prompt space. Given a frame feature vector ${\bf z}_f$, we seek a small number ($k$) of basis prompts that best approximate $h_{vid}({\bf z}_f)$. For this, we identify the $k$-dimensional subspace of $\cal P$ that has minimum least squares reconstruction error, i.e., the solution of 
\begin{eqnarray}
    \alpha_f^* = \arg\min_{\alpha_f, ||\alpha_f||_0 =k} \|h_{vid}({\bf z}_f) - \mathbf{F} \alpha_f\|_2,
   \label{eq:sparse}
\end{eqnarray} 
where $\mathbf{F} \in \mathbb{R}^{d_{f} \times B}$ contains the $B$ basis prompts in $\cal F$ as columns and $||\alpha||_0$ is the 0-norm (number of non-zero elements) of $\alpha$. We refer to this as the local reconstruction subspace for $h_{vid}({\bf z}_f)$  and denote the set of vectors
\begin{equation}
    \mathcal{S}_v^f = \{\mathbf{f}_i | \alpha_{f,i}^* \neq 0\}
    \label{eq:Svf}
\end{equation}
as the {\it best local reconstruction basis\/} for $h_{vid}({\bf z}_f)$. 

The vector $\alpha_{f}^*$ has a closed form solution due to the orthogonality of the basis $\cal F$. For any matrix $\mathbf{G}$ containing $k$ columns of $\bf F$, the minimizer of $\|{\bf h} - \mathbf{G} \alpha\|_2$ is 
\begin{equation}
    \alpha = ({\bf G}^T{\bf G})^{-1}{\bf G}^T {\bf h} = {\bf G}^T h_{vid}({\bf z}_f), \label{eq:alpha}
\end{equation}
since ${\bf G}$ is orthogonal, leading to the subspace distance 
\begin{eqnarray}
   &\|{\bf h} - \mathbf{G} \alpha\|_2 = ({\bf h} - \mathbf{G} \alpha)^T({\bf h} - \mathbf{G} \alpha)~~~~~~~~~~~~~~~~~~~~~~~~~~~~~~~~~~ \nonumber\\
   &~~~~~~~~~~~~~~~~~~~~~~~~~= \|{\bf h}\|_2 - 2 {\bf h}^T {\bf G}\alpha + \| \alpha \|_2 = \|{\bf h}\|_2 - \| \alpha \|_2.
\end{eqnarray}
It follows that the solution of Eq.~(\ref{eq:sparse}) is the subspace that maximizes the magnitude of the $\alpha$ vector. Since $\alpha$ has the form of Eq. (\ref{eq:alpha}), this occurs when $\alpha$ includes the largest $k$ dot-products $h_{vid}({\bf z}_f)^T {\bf f}_i$, i.e
\begin{equation}
    \alpha_{f,i}^* = 
        h_{vid}({\bf z}_f)^T {\bf f}_i \times \mathbbm{1}_{i \in s({\bf z}_f,{\cal F}; k, h_{vid})}
\end{equation}
where $\mathbbm{1}_{(\cdot)}$ is the indicator function,
\begin{equation}
    s({\bf z},{\cal F}; k, h) = \text{top-}k(\{{h({\bf z})}^T {\bf f}_1,...,{h({\bf z})}^T {\bf f}_B\})\label{eq:topk}
\end{equation}
and $\text{top-}k$ returns the indices of the largest $k$ elements of its argument. Given $\alpha_{f}^*$, the $k$ basis-prompts in the  best reconstruction basis $\mathcal{S}_v^f$ of Eq.~(\ref{eq:Svf}) are mapped to feature space by a latent space decoder $g_{vid}(\cdot): \mathbb{R}^{d_f} \to \mathbb{R}^{d_{vid}}$. This produces a set of prompts 
\begin{equation}
    \mathbf{P}_v^f = g_{vid}(\mathbf{F} {\bf A}_v^f)
    \label{eq:pv}
\end{equation}
where ${\bf A}_v^f \in \{0,1\}^{B \times k}$ is a matrix whose $i^{th}$ column is the one-hot code for the $i^{th}$ index in $s({\bf z}_f,{\cal F}; k, h_{vid})$.

\noindent\textbf{Prompt learning.} The learning goal is to derive the basis $\cal F$ of the prompt space over the entire dataset. This is the set of vectors $\mathbf{f}_i$ that minimize the reconstruction error of Eq.~(\ref{eq:sparse}) under the orthogonality constraint ${\bf f}_i^T {\bf f}_{j} = \mathbbm{1}_{i=j}$. We ensure that all vectors have unit norm by introducing a normalization layer after the encoder $h_{vid}(\cdot)$ and learn the basis by minimizing the Lagrangian 
\begin{equation}
    \sum_{f=1}^T\|\sum_{{\bf f}_i\in \mathcal{S}_v^f} \alpha_{f,i}^* {\bf f}_i - h_{vid}({\bf z}_f)\|_2 
    +\sum_{{\bf f}_i,{\bf f}_j\in {\cal F}} \xi_i {\bf f}_i^T {\bf f}_{j\neq i},\label{eq:rec_loss}
\end{equation}
where $\xi_i$ are Lagrange multipliers set to $\xi_i =1, \forall i$ in all experiments. 
Since the basis is optimized on the entire dataset, the subspace spanned by $\cal F$ is a global low dimensional approximation to the space of prompts, akin to a principal component analysis (PCA) of components $\mathbf{f}_i$. Under this view, the approach can be seen as the computation of a localized PCA, which selects the $k$ principal components that best approximate $h_{vid}({\bf z}_f)$ to synthesize the prompts of Eq.~(\ref{eq:pv}).

\noindent\textbf{Cross-frame context modeling.}
Since the frame-specific visual prompts ${\bf P}_v^f$ are conditioned on the frame feature ${\bf z}_f$, they encapsulate the context of frame $f$. Hence, cross-attention between ${\bf z}_{[CLS]}$ and the prompts ${\bf P}_v^1,...,{\bf P}_v^T$ can summarize knowledge across frames, without requiring additional modules like the bi-directional LSTM of CMM~\cite{huang2023vop}. The hyper-parameters $d_f$, $B$, and $k$ trade-off the number of parameters required to store the prompt basis, with the reconstruction error of the local least squares approximation, and the desired number of prompts to be added per transformer stage.  In any case, because the basis ${\cal F}$ and the encoder/decoder pair $h_{vid},  g_{vid}$, are the only learned parameters, the approach is very efficient. In all our experiments, both $h_{vid}$ and $g_{vid}$ are implemented with a single linear layer of $d_{vid} \times d_f$ parameters. Hence, the proposed approach only requires $d_f (B+2d_{vid})$ parameters, which is usually much smaller than the $(16+2M_vT)d_{vid}^2$ additional parameters of CMM. Please refer to the appendix for a detailed comparison.  In Table~\ref{table:main}, we show that good adaptation performance can be obtained with only $0.84\%$ of the VFM parameters.

\cutsubsection
\subsection{Cross-modal Prompt Synthesis}
\begin{figure}[t!]
    \centering
        \includegraphics[width=0.85\linewidth]{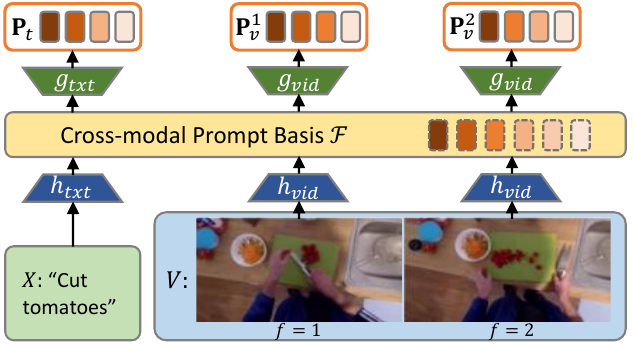}
        \vspace{-8pt}
        \caption{{\bf Cross-modal Prompt Synthesis.} The basis prompts $\cal F$ are shared across frames and modalities, but different mapping functions $h,g$ are adopted per modality to synthesize the prompts.
        }
        \label{fig:xps}
\end{figure}

\cutsubsection
While the text description and image frames originate from two modalities, the underlying semantic context should be shared, since both refer to the same video event (e.g. ``cut a tomato"). Intuitively, visual content in the video can benefit text domain features and vice versa.
To enable knowledge transfer between the two modalities, we propose cross-modal prompt synthesis, where the prompt basis ${\cal F}$ of section~\ref{sec:ps} is shared across modalities, as shown in Figure~\ref{fig:xps}.  

Similar to video prompt synthesis, a text feature ${\bf z}_t={\bf z}_{[EOS]}$ is mapped into prompt space by a text encoder $h_{txt}(\cdot): \mathbb{R}^{d_{txt}} \to \mathbb{R}^{d_f}$, $h_{txt}({\bf z}_t)$ is used to query the prompt basis ${\cal F}$ using $s({\bf z}_t,{\cal F}; k, h_{txt})$ in Eq.~(\ref{eq:topk}), and a prompt set produced with $\mathbf{P}_t = g_{txt}(\mathbf{F} {\bf A}_t)$, 
where ${\bf A}_t \in \{0,1\}^{B \times k}$ is a matrix whose $i^{th}$ column is the one-hot code for the $i^{th}$ index in $s({\bf z}_t,{\cal F}; k, h_{txt})$. $g_{txt}(\cdot): \mathbb{R}^{d_f} \to \mathbb{R}^{d_{txt}}$ is an additional prompt generator that maps basis vectors into text prompts.
The projection functions $h_{vid}$, $h_{txt}$ and joint basis ${\cal F}$ are jointly optimized with a loss
\begin{equation}
    {\cal L}_{syn} = \frac{1}{|\cal B|}\sum_{i\in\mathcal{B}} l_s(V_i, X_i)\label{eq:syn_loss}
\end{equation}
where
\begin{eqnarray}
    &l_s(V, X) = \sum_{f=1}^T\|\sum_{{\bf f}_i\in \mathcal{S}_v^f} \alpha_{f,i}^* {\bf f}_i - h_{vid}({\bf z}_f)\|_2~~~~~~ \nonumber \\
   &+ \|\sum_{{\bf f}_i\in \mathcal{S}_t} \alpha_{t,i}^* {\bf f}_i - h_{txt}({\bf z}_t)\|_2 + 
    \sum_{{\bf f}_i,{\bf f}_j\in {\cal F}} {\bf f}_i^T {\bf f}_{j\neq i}.
\end{eqnarray}
with $\alpha_{t,i}^* = h_{txt}({\bf z}_t)^T {\bf f}_i \times \mathbbm{1}_{i \in s({\bf z}_t,{\cal F}; k, h_{txt})}$, similar to Eq.~(\ref{eq:rec_loss}).

\cutsubsection
\subsection{Training}\label{sec:training}
\cutsubsection
The prompt basis ${\cal F}$ and projections $h_{vis}$, $h_{txt}$, $g_{vis}$ and $g_{txt}$, are jointly optimized using a combination 
\begin{equation}
    {\cal L} = {\cal L}_{cl} + \lambda {\cal L}_{syn}, 
    \label{eq:all_loss}
\end{equation}
of the contrastive loss of Eq.~(\ref{eq:nce_loss}) and the cross-modal prompt synthesis loss of Eq.~(\ref{eq:syn_loss}), where $\lambda$ is a hyperparameter. However, since the prompt basis is randomly initialized, the basis prompts do not contain semantic information in the early stages of training. This problem is compounded by the use of top-$k$   basis prompt selection, where only the basis prompts will be updated, and these updated basis prompts will then be selected again in the next iteration.  To prevent this, instead of using the top-$k$ selection rule during training, we sample $k$ basis prompts from a multinomial distribution $\pi_m$, which is a mixture 
\begin{equation}
    \pi_m = \gamma \pi_{sim} + (1-\gamma) \pi_{invf},\label{eq:sampling}    
\end{equation}
of the distribution $\pi_{sim}$ of similarities between query and basis prompts, and the inverse of the basis prompt selection frequency $\pi_{invf}$. The mixture coefficient $\gamma$ is set to 0 at the beginning of training and gradually increased to 1 in later epochs. This increases the possibility that basis prompts rarely seen in the dataset are learned. Note that the top-$k$ basis prompts are always selected during inference.

\cutsectionup
\section{Experiments}
\cutsectiondown
\begin{table}[t!]
    \centering
        \setlength{\tabcolsep}{8pt}
        \resizebox{\linewidth}{!}{
        \begin{tabular}{l|c|c|cc}
        \toprule
        \multirow{2}{*}{Method} & Tunable  & Charades-Ego & \multicolumn{2}{c}{EGTEA} \\ \cline{3-5}
          & Params (\%) & mAP & Mean Acc & Top-1 Acc\\ \midrule
          Zero-shot &    $0\%$       & $26.8$ & $28.90$ & $35.51$\\
          Full fine-tuning     &   $100\%$       & $32.9/33.7^{\star}$ & $67.77$ & $71.37$ \\ \midrule \midrule
          Bias~\cite{cai2020tinytl,zaken2021bitfit}      &   $0.12\%$      & $30.0$  & $55.29$ & $61.52$ \\
          TPT~\cite{coop}          &   $0.002\%$   & $29.7$ & $51.93$ & $58.36$ \\
          VPT~\cite{jia2022visual} &   $0.66\%$   & $31.7$ & $63.11$ & $68.35$\\
          VoP~\cite{huang2023vop}  & $0.67\%$ & $32.5$ & $66.36$ & $70.72$\\
          VoP\textsuperscript{C}~\cite{huang2023vop}          & $10.64\%$ & $32.4$ & $67.55$ & $71.91$\\
          VoP\textsuperscript{F+C}~\cite{huang2023vop}      & $10.86\%$ & $32.7$ & $68.70$ & $73.24$\\ \midrule
          Ego-VPA (Ours)    & $0.84\%$  & $\bf 33.8$ & $\bf 69.17$  & $\bf 73.39$\\  \bottomrule
        \end{tabular}
        }
        \vspace{-7pt}
        \captionof{table}{Results on Charades-Ego and EGTEA compared to state-of-the-art prompt-tuning methods introduced in section~\ref{sec:settings}. 
        $^{\star}$ denotes the number in~\cite{zhao2023lavila}, using $4\times$ batch size compared to ours.}\label{table:main}
\end{table}
In this section, we validate the efficiency and effectiveness of Ego-VPA by comparing it with SOTA methods and ablating on different components. More results in the appendix.
\cutsubsection
\subsection{Experimental Setup}\label{sec:exp_setup}
\cutsubsection
\noindent\textbf{Datasets.}
We adopt LaViLa~\cite{zhao2023lavila} as the Ego-VFM, which is pretrained on Ego4D~\cite{grauman2022ego4d}, containing 4M video-text pairs for egocentric videos. Both the proposed method and baselines are evaluated on Charades-Ego~\cite{sigurdsson2018charades}, EGTEA~\cite{li2018eye}, and EPIC-Kitchens-100~\cite{damen2022rescaling}, which are widely used in egocentric video research. Charades-Ego~\cite{sigurdsson2018charades} is a fine-grained action classification dataset containing $33,114$ trimmed action segments for training, spanning across 157 classes. While both egocentric and exocentric views are provided, we only train and evaluate egocentric videos using the official splits, as in ~\cite{kevin2022egovlp,zhao2023lavila}. EGTEA~\cite{li2018eye} is an egocentric cooking video dataset, containing $10,321$ action instances from 106 fine-grained classes. We only use the video data and follow the same protocol for training and evaluation as in~\cite{zhao2023lavila}. EPIC-Kitchens-100~\cite{damen2022rescaling} is an egocentric cooking video dataset of $100$ hours, containing $67,217$/$9,668$ clips for training/validation. We evaluate the multi-instance retrieval (MIR) task to test the generalization of Ego-VPA, which contains a text-to-video (T->V) and video-to-text (V->T) retrieval task.

\noindent\textbf{Metrics.}
At inference, video and text features are extracted with $\phi_{vid}$ and $\phi_{txt}$ respectively. For action classification, we compute the cosine similarity per video between the video feature and the text feature of each action class as the classification score. We report commonly used metrics for these datasets. Since each testing video in Charades-Ego is multi-label, we report the mean average precision (mAP) over the 157 classes.
For EGTEA, we report top-1 accuracy (Top-1 Acc) and average accuracy over all classes (Mean Acc). For the EPIC-Kitchens-100 MIR task, mean Average Precision (mAP) and Normalized Discounted Cumulative Gain (nDCG) for T->V and V->T retrieval are reported.

\noindent\textbf{Model architecture.}
The Ego-VFM architecture is inherited from LaViLa~\cite{zhao2023lavila}, where the text encoder $\phi_{txt}$ is a 12-layer ($L=12$) Transformers with $d_{txt}=512$ and the video encoder $\phi_{vid}$ a 12-layer ($L=12$) TimeSformer with $d_{vid}=768$. We prompt the model with $M_v=M_t=8$, $K=8$, setting $d_f=512, B=10$ for the prompt basis. All models are trained with $16$ frames per video ($T=16$) unless explicitly noted.
More details on implementation are reported in the appendix.

\noindent\textbf{Computing.}
As in section~\ref{sec:settings}, we freeze the Ego-VFM model parameters and only learn a small portion of prompts on the downstream datasets. Ego-VPA does not require large computing resources compared to prior work. We train all the experiments with a batch size of 4 per GPU using 8 NVIDIA Titan Xp GPUs. This is $4\times$ smaller than the batch size in~\cite{zhao2023lavila}. The memory required for Ego-VPA training is around $2/3$ of that required to fine-tune the full model.

\begin{table}[t!]
    \centering
    \setlength{\tabcolsep}{9pt}
    \resizebox{\linewidth}{!}{
    \begin{tabular}{l|cccc|c}
        \toprule
         & Prompt & Cross & Orthogonality & Prompt & \multirow{2}{*}{mAP}\\ 
         & Generation & Modality & Constraint & Query & \\ \midrule
         (m1) & CMM &  & N/A & N/A &  32.7\\ \midrule
         (m2) & PS &  & & $\sim\pi_m$ &  32.8 \\
         (m3) & PS &  & \checkmark & $\sim\pi_m$ & 33.0 \\
         (m4) & PS & \checkmark &  & $\sim\pi_m$ &  33.3 \\
         (m5) & PS & \checkmark & \checkmark & $\sim\pi_m$ & {\bf 33.8} \\
         (m6) & PS & \checkmark &  & top-$k$ & 32.8 \\
         (m7) & PS & \checkmark & \checkmark & top-$k$ & 33.5\\
         \bottomrule
    \end{tabular}
    }
    \vspace{-7pt}
    \caption{Ablations on prompt generation, the orthogonality constraint imposed by the $2^{nd}$ term of Eq.~(\ref{eq:rec_loss}), and the prompt query (PS: prompt synthesis; $\sim\pi_m$: sampling Eq.~(\ref{eq:sampling})). (m1) is VoP\textsuperscript{F+C}~\cite{huang2023vop}; (m2-7) are Ego-VPA variants, and (m5) is full Ego-VPA.}
    \label{table:ablation}
\end{table}

\cutsectionup
\subsection{Comparisons to SOTA Prompt-tuning Methods}\label{sec:sota}
\cutsectiondown
Table~\ref{table:main} presents the result on Charades-Ego and EGTEA, as a function of the tunable model parameters. The poor zero-shot performance shows that current Ego-VFMs, like LaViLa, are somewhat overfitted to the Ego4D dataset. Full fine-tuning significantly improves performance, but requires optimization of the entire VFM, which is inefficient. Prompt-tuning methods require less parameter optimization.
Bias~\cite{cai2020tinytl,zaken2021bitfit} only fine-tunes the bias terms in the model but not the weights, and thus has limited adaptation capacity.
TPT~\cite{coop} and VPT~\cite{jia2022visual} extend the input sequence with learnable prompts for the text and video encoders, respectively. These methods are weaker than VoP~\cite{huang2023vop}, where prompt-tuning is performed for both encoders. 
VoP\textsuperscript{F+C}, a variant of VoP using a CMM module and frame-aware attention layers, is the best-performing baseline. However, the introduction of the CMM module significantly increases the parameter counts, requiring around $10\%$ of the model size for the adaptation. The proposed Ego-VPA consistently outperforms all other methods on both datasets, even beating full fine-tuning, with only $0.84\%$ trainable parameters.
While the adaptation to EGTEA produces much stronger results, indicating that the domain gap to Ego4D is smaller, the zero-shot performance of LaViLa is still quite weak, reinforcing the importance of efficient adaptation methods. 

\begin{figure*}[t!]
    \centering
    \resizebox{0.95\linewidth}{!}{
    \centering
    \begin{subfigure}[t]{0.24\linewidth}
        \centering
        \hspace{-10pt}
        \includegraphics[width=\linewidth]{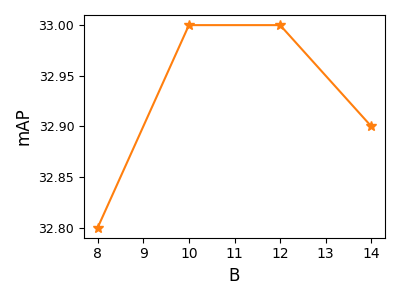}
        \vspace{-5pt}
        \caption{\footnotesize}\label{fig:B}
    \end{subfigure}
    \begin{subfigure}[t]{0.24\linewidth}
        \centering
        \hspace{-10pt}
        \includegraphics[width=\linewidth]{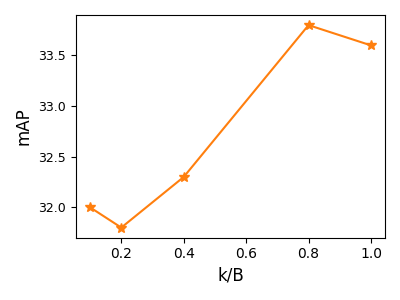}
        \vspace{-5pt}
        \caption{\footnotesize}\label{fig:k}
    \end{subfigure}
    \begin{subfigure}[t]{0.24\linewidth}
        \centering
        \hspace{-10pt}
        \includegraphics[width=\linewidth]{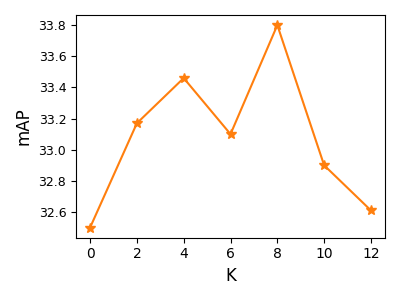}
        \vspace{-5pt}
        \caption{\footnotesize}\label{fig:K}
    \end{subfigure}
    \begin{subfigure}[t]{0.24\linewidth}
        \centering
        \hspace{-10pt}
        \includegraphics[width=\linewidth]{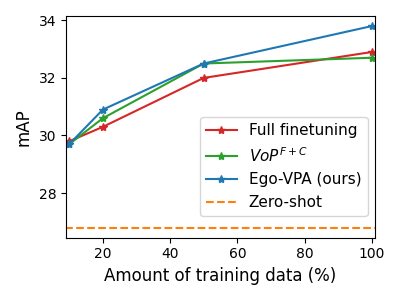}
        \vspace{-5pt}
        \caption{\footnotesize}\label{fig:lowdata}
    \end{subfigure}
    }
    \vspace{-12pt}
    \caption{(a) Ablations on the size of prompt basis $B$ (using $k=8$). (b) Ablations on the size of the reconstruction basis $k$ given a fixed $B$. (c) Ablations on {\it intra/inter-frame} attention boundary $K$.  (d) Ablations on different amounts of training data.}
\end{figure*}

\cutsectionup
\subsection{Ablation Studies}
\cutsectiondown
This section ablates different designs of Ego-VPA with the Charades-Ego dataset, unless explicitly noted.

\noindent\textbf{Context modeling module.}
We first compare the proposed prompt synthesis (PS) with the CMM module of $VoP^{F+C}$~\cite{huang2023vop}. As shown in Table~\ref{table:ablation}, despite using much fewer trainable parameters, video-only prompt synthesis (m3) outperforms the CMM in $VoP^{F+C}$ (m1).  When cross-modal knowledge transfer is enabled with cross-modal prompt synthesis (m5), the performance further improves largely. This validates the effectiveness of context modeling across frames and modalities.

\noindent\textbf{Orthogonality constraint.}
The closed-from solution of Eq. (\ref{eq:alpha}) only holds if the basis-prompts in $\cal F$ are orthogonal. The orthogonality constraint in the $2^{nd}$ term of Eq.~(\ref{eq:rec_loss}) is important to guarantee this property. This is validated by (m2-3), (m4-5), and (m6-7) in Table~\ref{table:ablation}, where a gain is consistently observed when the orthogonality constraint is imposed, especially in the cross-modal setting.

\noindent\textbf{Prompt query strategies.}
Table~\ref{table:ablation} compares the pure top-$k$ selection rule with the sampling strategy of Eq.~(\ref{eq:sampling}) (m4-7).
The latter encourages every basis prompt in ${\cal F}$ to be updated in the early training stages. In the later stages, as $\gamma$ approaches 1, it reduces to top-$k$ selection. (m4-7) shows that the strategy of Eq.~(\ref{eq:sampling}) allows more distributed feature updates and improves performance.

\noindent\textbf{Size of prompt basis and reconstruction basis.}
In Figure~\ref{fig:B}, we ablate the size of prompt basis $B$ when $k=8$ for video prompt synthesis.
Note that we use 8 visual prompts per-frame (i.e. $M_v=8$), so $B=8$ means that all basis prompts are used to generate prompts.
Increasing $B$ allows a slight improvement since some features may not co-exist across frames. However, the performance saturates quickly, suggesting that the prompt space has a very low dimension. Conversely, we ablate the size of reconstruction basis $k$ given a fixed $B$ for cross-modal prompt synthesis in Figure~\ref{fig:k}. In general, increasing $k$ enhances the expressiveness of the projected features, since more basis prompts are used for feature recovery. However, there is a benefit to the local subspace approximation, as best results are usually obtained for $k/B < 1$. In Figure~\ref{fig:k}, the optimal ratio is $0.8$.

\noindent\textbf{Intra/inter-frame attention boundary.}
As discussed in section~\ref{sec:ps}, we adopt {\it intra-frame} attention in the first $K$ layers of the video encoder and {\it inter-frame} attention in the remaining, allowing shallow layers to adapt lower-level features, and deeper layers to fuse high-level semantics.
We ablate such attention boundary by training the model with different values of $K$, as shown in Figure~\ref{fig:K}. Note that both visual and text encoders have 12 layers (i.e. $L=12$). Results show that setting $K=8$ leads to the best performance, coherent to the observation of~\cite{huang2023vop}.

\noindent\textbf{Number of frames.}
We validate the robustness of Ego-VPA by training with different numbers of frames
on both Charades-Ego and EGTEA. As shown in Table~\ref{tab:frame}, the performance of all methods increases with the number of frames, indicating that temporal resolution is an important factor for video understanding. Ego-VPA consistently improves over $VoP^{F+C}$ and the full fine-tuning performances, showing that it is robust and effective across time resolutions.

\noindent\textbf{Amount of training data.}
To evaluate the proposed Ego-VPA on the low-data regime,
we adapt the models with $10\%$, $20\%$, $50\%$, and $100\%$ data respectively. As shown in Figure~\ref{fig:lowdata}, both Ego-VPA and the SOTA VoP\textsuperscript{F+C} are effective for various amounts of training data, reaching comparable or even superior performance than full fine-tuning and improving largely over the zero-shot results. However, the SOTA model still underperforms Ego-VPA in general.

\begin{table}[t!]
    \centering
        \setlength{\tabcolsep}{9pt}
        \resizebox{\linewidth}{!}{
        \begin{tabular}{l|ccc|ccc}
            \toprule
             Dataset & \multicolumn{3}{c|}{Charades-Ego (mAP)} & \multicolumn{3}{c}{EGTEA (mAcc)} \\ \midrule
             \# of frames & 4 & 8 & 16 & 4 & 8 & 16 \\ \midrule
             Zero-shot & 24.4 & 26.0 & 26.8 & 27.0 & 28.5 & 28.9 \\
             Full fine-tuning & 28.3 & 31.4 & 32.9 & 56.1 & 62.9 & 67.8\\ \midrule\midrule
             VoP\textsuperscript{F+C}~\cite{huang2023vop} & 28.6 & 30.9 & 32.7 & 59.1 & 63.6 & 68.7\\
             Ego-VPA (ours) & $\bf 29.3$ & $\bf 31.5$ & $\bf 33.8$ & $\bf 60.5$ & $\bf 64.4$ & $\bf 69.2$\\
             \bottomrule
        \end{tabular}
        }
        \vspace{-5pt}
        \captionof{table}{Ablations on using different numbers of frames per video.}
        \label{tab:frame}
\end{table}

\begin{figure}[t!]
        \centering
        \includegraphics[width=0.95\linewidth]{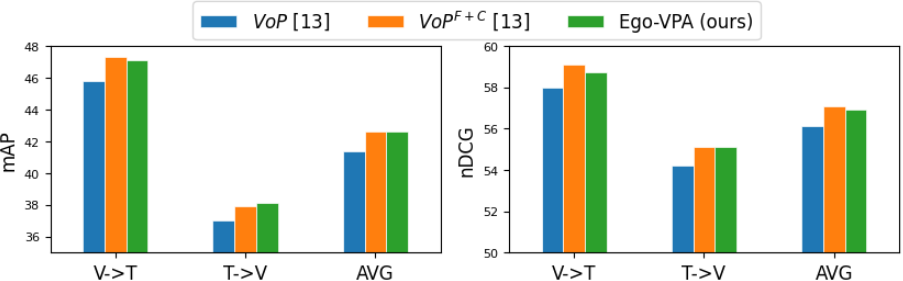}
        \vspace{-10pt}
        \captionof{figure}{Ego-VPA can generalize to EPIC-Kitchens-100 multi-instance retrieval task.}
        \label{fig:ek100}
\end{figure}

\cutsectionup
\subsection{Generalization to Retrieval Tasks}
\cutsectiondown
We further evaluate multi-instance retrieval tasks~\cite{damen2022rescaling} on Epic-Kitchens-100. 
Figure~\ref{fig:ek100} shows that Ego-VPA performs on par with $VoP^{F+C}$, which has $10\%$ more trainable parameters. Compared to vanilla VoP, which has a similar number of parameters, Ego-VPA has clearly better performance. This shows that Ego-VPA is more parameter-efficient and effective across different egocentric video tasks.

\cutsectionup
\section{Conclusions}
\cutsectiondown
We propose Ego-VPA, a novel parameter-efficient adaptation method for Ego-VFMs. Atop a frozen Ego-VFM, we sparsely approximate projected video frame/text features with a shared prompt basis and synthesize video/text prompts accordingly. This is shown to enhance context fusion across frames and cross-modal transfer, achieving improved visual-language alignments. Through extensive experiments, we show that Ego-VPA is both efficient and effective, outperforming SOTA methods with much fewer learnable parameters.

\noindent\textbf{Acknowledgment}
This work was partially funded by NSF awards IIS-2303153, and a gift from Qualcomm.

{\small
\bibliographystyle{ieee_fullname}

}

\end{document}